\documentclass[preprint,12pt,authoryear]{elsarticle}

\usepackage{hyperref}
\usepackage{booktabs}
\usepackage{multirow}
\usepackage{amssymb}
\usepackage{booktabs}
\usepackage{graphicx}%
\usepackage{multirow}%
\usepackage{amsmath,amssymb,amsfonts}%
\usepackage{amsthm}%
\usepackage{mathrsfs}%
\usepackage[title]{appendix}%
\usepackage{xcolor}%
\usepackage{textcomp}%
\usepackage{manyfoot}%
\usepackage{booktabs}%
\usepackage{algorithm}%
\usepackage{algorithmicx}%
\usepackage{algpseudocode}%
\usepackage{listings}%
\usepackage{tabularx}
\usepackage{multirow}
\usepackage{amsthm}
\usepackage{hyperref} 

\newtheorem{definition}{Definition}

\begin{document}

\begin{frontmatter}


 \title{KGroot: Enhancing Root Cause Analysis through Knowledge Graphs and  Graph Convolutional Neural Networks}
 \author[1]{Tingting Wang} 
 \ead{tingtingwang@seu.edu.cn}
 \author[1]{Guilin Qi\corref{cor1}}
 \ead{gqi@seu.edu.cn}
 \author[1]{Tianxing Wu}
  \ead{tianxingwu@seu.edu.cn}
 \cortext[cor1]{Corresponding author}
 \affiliation[1]{organization={School of Computer Science and Engineering, Southeast University},city={ Nanjing},country={P.R. China}}







\begin{abstract}
Fault localization is challenging in online micro-service due to the wide variety of monitoring data volume, types, events and complex interdependencies in service and components.  Faults events in services are propagative and can trigger a cascade of alerts in a short period of time.  In the industry, fault localization is typically conducted manually by experienced personnel. This reliance on experience is unreliable and lacks automation. Different modules present information barriers during manual localization, making it difficult to quickly align during urgent faults. This inefficiency lags stability assurance to minimize fault detection and repair time. Though actionable methods aimed to automatic the process, the accuracy and efficiency are less than satisfactory. The precision of fault localization results is of paramount importance as it underpins engineers trust in the diagnostic conclusions, which are derived from multiple perspectives and offer comprehensive insights. Therefore, a more reliable method is required to automatically identify the associative relationships among fault events and propagation path. To achieve this, KGroot uses event knowledge and the correlation between events to perform root cause reasoning by integrating knowledge graphs and GCNs for RCA. FEKG is built based on historical data, an online graph is constructed in real-time when a failure event occurs, and the similarity between each knowledge graph and online graph is compared using GCNs to pinpoint the fault type through a ranking strategy. Comprehensive experiments demonstrate KGroot can locate the root cause with accuracy of 93.5\% top 3 potential causes in second-level. This performance matches the level of real-time fault diagnosis in the industrial environment and significantly surpasses state-of-the-art baselines in RCA in terms of effectiveness and efficiency. (KGroot is available at \url{https://github.com/daixixiwang/KGroot})
\end{abstract}



\begin{keyword}
Root Cause Analysis\sep Faults Locating\sep  Knowledge Graph\sep GCN


\end{keyword}

\end{frontmatter}


\section{Introduction}
\label{intro}
According to the data from the 2023 Cloud Computing White Paper, the global cloud computing market reached 491 billion dollars and grew steadily at a rate of 19\% last year. It is expected that under the stimulation of demands for large models and computing power, the cloud computing industry will maintain steady growth, and the global cloud computing market will exceed one trillion dollars by 2026\citep{caict2023report}. With the progression of cloud-native technologies, containerization has become more prevalent, leading to an increased frequency of container scheduling. The interdependencies and interactions among microservices are intricate, forming a complex network of relationships. Moreover, cloud centers frequently encounter scenarios that challenge operational stability, such as capacity overload due to abrupt traffic surges, network fluctuations, host outages, dependency breakdowns, and database connection saturation. The complexity and dynamism of infrastructure and PaaS components can have a direct or indirect impact on service stability, which in turn is critical for meeting the Service Level Agreement (SLA) requirements of users. How to swiftly diagnose faults in complex environments so as to reduce Mean Time To Repair (MTTR) is a crucial aspect of Artificial Intelligence for IT Operations(AIOps).\\
\indent Though tremendous effort has been devoted to software service maintenance (e.g., various metrics, such as average response time or memory usage, are closely monitored on a 24×7 basis\citep{chen2022adaptive}), failures are inevitable due to the large scale and complexity, causing huge economic loss and user dissatisfaction\citep{li2022actionable}\citep{Tencent2023}\citep{aliyun2023notice}\citep{cheng2023ai}. \\
\indent \textbf{Monitoring Complexity}. An extensive array of data types, encompassing metrics, logs, traces, and various activities, necessitates comprehensive monitoring, processing, and storage. These activities may include service deployment, configuration modifications, or pod auto-scaling events. To effectively tackle root cause analysis(RCA), a substantial body of research has been produced. However, existing approaches mainly focus on unactionable fault levels, individual metrics\citep{yu2023cmdiagnostor}\citep{zhao2023multi} or components\citep{liu2021microhecl}\citep{gan2022practical}, which is difficult to meet the requirements of fault localization in microservices system. Systems dedicated to RCA ought to be adept at managing this heterogeneous and unstructured data landscape, with the capability to derive meaningful insights from the amassed information. \\
\indent \textbf{Dependencies Complexity}. Microservices architecture presents a labyrinth of interdependencies, with these services demonstrating a dependency on the complexities embedded within PaaS ecosystems. To be actionable, what occurs and which kind the fault is are two key problems in RCA. Anomalies, when they occur, can trigger a domino effect, leading to cascading failures that propagate through multiple services. Faults events are nature of propagative. In such scenarios, it is imperative for the RCA system to assimilate a diverse array of anomalous indications emanating from a plethora of alerts, identify the initial cause among numerous abnormal alarms and to discern the chronological and causal sequence of events.\\
\indent \textbf{Inefficiency and Unpromising Accuracy}. The knowledge gap between the centered SRE team and the domain SRE team gets further enlarged and makes RCA much more challenging\citep{yeruva2023aiops}\citep{kumar2023aiops}. Manual analysis duration exceeds the minute threshold, while expeditious response and swift pinpointing of issues require a minimum of 3-5 minutes, a timeframe which falls short of satisfactory standards. 
Despite the existence of several studies on automatic Root Cause Analysis (RCA),  their performance is less than satisfactory as they neglect the events and the associative relationships among them. Accuracy \citep{rudin2022interpretable} is crucial for SREs to have confidence in fault localization outcomes as it offers explanations from various viewpoints.\\
\indent \citep{dogga2022revelio} find that recurring categories of root causes cause 94\% of the failures at a major SaaS company. \citep{li2022actionable} categorizes root causes of failures into recurring and non-recurring by judging failures' repetitive with historical failures. Some recurring failures are mainly external, hardware, such as bad requests, disk I/O error, slow databased queries, and third-party service faults. While the non-recurring categories are mainly code logical reasons, such as bugs, error configuration. \\
\indent Motivated by the recurrence of most failures, where each fault is event-driven, with the aim of efficiently and effectively diagnosing recurring failures from historical data and online events in complex, dependent microservices, we propose a novel approach, KGroot,  leveraging knowledge graphs and graph convolutional neural networks(GCNs) to automatic RCA in complex micro-services. To tackle monitoring and dependencies complexity, we choose metrics from historical failures include component and service operational indicators, as logs are of great volume, lacking aggregation and index information, traces are of low sampling rate and may be incomplete chains for some reasons, such as outdated client version. To address the issues of inefficiency and unpromising accuracy, based on the recurring nature of faults, KGroot defines FPGs for each failure events according to historical failures, then, constructs FEKGs over FPGs for different scenario. KGroot builds FPGs and FEKGs  based on events. Event-driven RCA are of fine granularity, as events are based on monitoring metrics, which makes event-graph consistent with monitoring granularity. Such knowledge graphs consist of event types and causal relations. When an online failure recurs, KGroot constructs FPG dynamically according to online event graph, and computes graph similarity of online FPG with FEKGs through GCN. Finally, the top N most similar knowledge graph’s root cause type is determined. Depending on this root cause type, the most similar failure scenarios root cause event can be found by a ranking strategy. 
This paper's primary contributions can be summarized as follows:
\begin{itemize}
  \item We have for the first time proposed an event-driven knowledge graph to address RCA frequently occurring under complex dependencies and propagation relationships. 
  \item We propose an event knowledge graph based localization model to tackle the complexity of root cause analysis in microservice systems. 
  \item Comprehensive experiments are conducted on real and productive datasets and results verify KGroot significantly outperforms state-of-the-art baselines both in effectiveness and efficiency,matching the level of real-time fault diagnosis in the industrial environment. 
\end{itemize}
\indent The remainder of this paper is organized as follows. Section 2 reviews related work on the RCA problem, discussing methods that are based on probability, topology, causality, and graph-based techniques. Section 3 provides a high-level overview of the proposed KGroot model, outlining its workflow. Section 4 details the implementation of KGroot for RCA, addressing the construction of the fault propagation graph, the establishment of the fault event knowledge graph, and presenting the KGroot RCA methodology comprehensively. Section 5 presents the experimental results, evaluating KGroot against seven baseline models and two degraded versions in terms of both effectiveness and efficiency. Finally, Sections 6 concludes the paper with a discussion of the proposed method and directions for future research, respectively.


\section{Related work}\label{Related work}
Several comprehensive surveys exist that categorize previous research into distinct categories: probability-based approaches, topology graph, causality graph, ML-based approaches, and graph-based approaches. The following is an introduction to some of the representative approaches.\\
\indent \textbf{Probability-based RCA}. A traditional model for RCA, probability-based model is capable of performing analysis automatically and dynamically. \citep{chen2023balance} analyzed and studied the probability model based on Bayesian network. Despite being the most commonly used approach for RCA, it proves challenging to implement in practice due to the fact that Bayesian inference is an NP-hard problem.\\
\indent \textbf{Topology Graph RCA}. Brandn et al.\citep{brandon2020graph} constructs a system topology graph from the combination of Key Performance Indicators(KPIs) metrics and log data and extracts anomalous subgraphs from it and compares them with the normal system graph to detect the root cause. This approach constructs a pattern library which contains the anomalous device subgraph with labeled root cause nodes, which is difficult to determine the scale of the anomalous device subgraph. CloudRCA\citep{zhang2021cloudrca} uses a knowledge informed Hierarchical Bayesian Network over features extracted from metric and log-based anomaly detection to infer the root causes.  CloudRCA makes use of heterogeneous multi-source data including KPIs, logs, as well as topology, and extracts important features. And constructs Knowledge-informed Hierarchical Bayesian Network(KHBN) to infer root causes in Alibaba company. \\
\indent \textbf{Causality Graph RCA}. Researchers\citep{wang2023interdependent} use causality inference on KPI metrics and service call graphs to localize root causes in micro-service systems. Groot\citep{wang2021Groot} from eBay company, constructs a real-time causality graph based on events that summarize various types of metrics, logs, and activities in the system under analysis in industrial environment. \citep{wang2023incremental} develop an online trigger point detection approach based on multivariate singular spectrum analysis and cumulative sum statistics.\\
\indent \textbf{ML-Based RCA}. ML algorithms Decision Tree\citep{zhou2021machine}, Gradient Boosting, Random Forest\citep{rigatti2017random} , SVM can automate and continuously learn, but may require a large amount of training data, and there may be challenges in the interpretability of the model. A recently supervised method  D\'ej\`aVu\citep{li2022actionable} can automate manual faults localization in good accuracy and efficiency, but there is still room for optimization and improvement in accuracy and efficiency.\\
\indent \textbf{Graph-based RCA}. Graph-based RCA methods utilize graphical representations due to graph potent ability to express complex data structures. By creating graphical structures, researchers can address the issues of dependencies and propagation in faults\citep{wang2023incremental}. \citep{qiu2020causality} constructs a knowledge graph over entities extracted from logs, metrics, traces and system dependency graphs and mines causal relations using PC algorithm to detect root causes of incidents. By extracting rules from random forests built on log data using heuristics and domain knowledge, a knowledge base is continuously built to diagnose and classify performance failure issues in online manner\citep{bansal2020decaf}. Researchers adopt graph similarity and matching approaches in RCA.  JSS\citep{brandon2020graph} detects anomalies from KPI metrics and log data. It sources failures as graphs and finds similar diagnosed failures by graph similarity for each incoming failure. iSQUAD\citep{ma2020diagnosing} clusters historical failures and assigns each incoming failure to a cluster according to graph similarity. Even though the aforementioned research approach is viable and comprehensible, it falls short when it comes to the accuracy and efficiency of predictions in RCA. \\
\indent Based on the openness and advancement of methods, and reproducibility of based data, we have chosen the state-of-the-art methods as the baseline for this paper. In this paper, an automatic RCA method leveraging knowledge graphs and GCN is proposed. Experimental results prove the effectiveness of proposed method than state-of-the-art methods.
 
\section{Overview}\label{overview}
\begin{figure}[h]%
\centering
\includegraphics[width=0.9\textwidth]{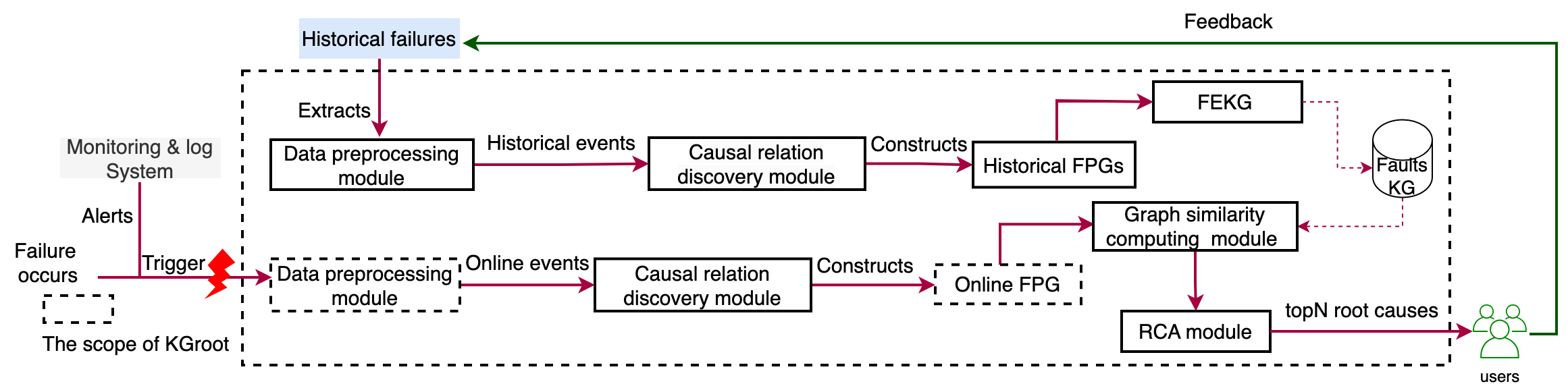}
\caption{Workflow of KGroot}\label{workflow of KGroot}
\end{figure} 

\begin{figure}[h]%
\centering
\includegraphics[width=0.9\textwidth]{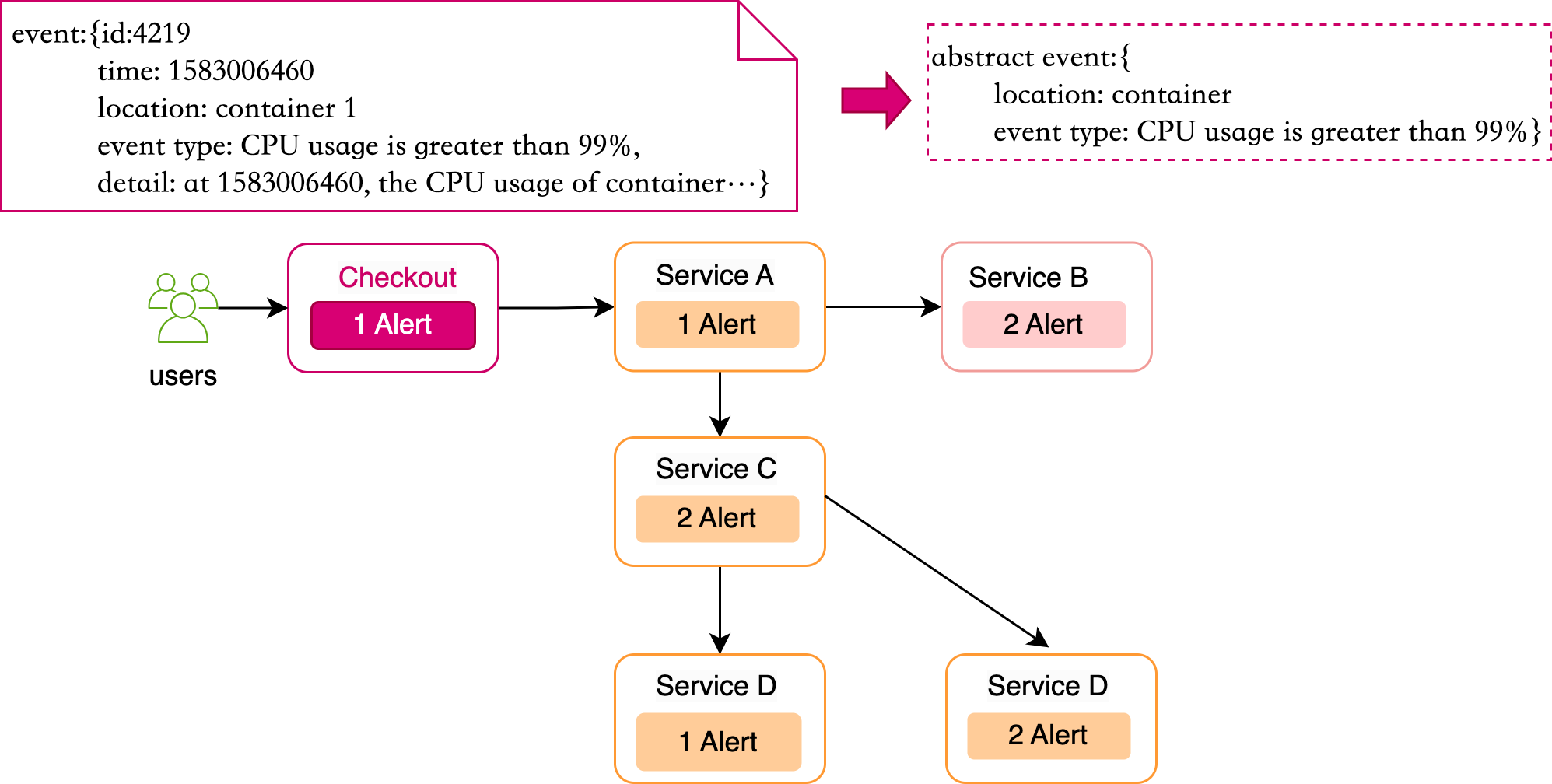}
\caption{An example of event transformed to abstract event in alerting}\label{event}
\end{figure}

KGroot comprises four key modules: the data preprocessing module, the causal relation discovery module, the graph similarity computation module, and the RCA module. Figure \ref{workflow of KGroot} illustrates the structure of our proposed KGroot. The data preprocessing module transforms logs and metrics unstructured data into structured events. Subsequently, the causal relation discovery module reveals the causal connections between these events, resulting in the creation of knowledge graphs and online graphs. The graph similarity computation module computes online graph similarity with each knowledge graphs. The RCA module then determines the fault's root cause by selecting the top N most similar knowledge graph as identified by the graph similarity computation module.\\
\indent Within the data preprocessing module, unstructured data, such as logs and metrics, are converted into structured events. Take, for instance, an event $e_1$ is transformed into structured abstract event $e'_1$ depicted as follows. Figure \ref{event} illustrates an example of event.

\indent The causal relation discovery module is to identify causal relationships between events, which are subsequently utilized to build the knowledge graph and online graph. The knowledge graph is constructed using historical data prior to the fault's occurrence, while the online graph is a real-time graph built when a fault arises. The creation of a knowledge graph, corresponding to specific root causes and fault types, integrates the automatically constructed knowledge graph with the expert's prior knowledge. The purpose of the online graph is to extract the most significant information from online fault data, thereby preventing the diagnosis from being disrupted by excessive noise information. The module employs the Support Vector Machine(SVM) to distinguish whether there's a causal relation between two events, aiming to recognize logically valuable relationships rather than simply treating them as event sequences.\\
\indent Within the graph similarity computation module, given that an online graph, containing the causal relations between events are essentially the same for the same root cause and fault type, this module identifies the knowledge graph that most closely resembles the online graph. Each knowledge graph is labeled the root cause of the fault by domain experts or administrators. The Graph Convolutional Network(GCN) is utilized to embed events with causal relations to compute the similarity of online graph and knowledge graph.\\
\indent In the real scenario, apart from reminding users of what happens in this system causing the fault, the location of root causes should also be pinpointed to help users diagnose and maintain systems. RCA module synthetically considerate time and distance between alarm event and candidate root cause events. Ultimately, the RCA module returns the top N most similar labeled knowledge graph to users, implying that both the root cause event type and fault type have been identified.

\section{KGroot RCA}\label{KGroot}
\subsection{FPG Construction}\label{FPG}
In KGroot, it's necessary to create failure propagation graphs both based on the accumulated knowledge from historical failures and the online fault at the time of a fault occurrence. The following is the definition of the online fault propagation graph.

\begin{definition}Fault Propagation Graph(FPG).
FPG is a collection of triples $(e_{1}, r, e_{2})$, where $e_{1}$ and $e_{2}$ are events, and r is the relationship between events $e_{1}$ and $e_{2}$. $r$ is either a sequential relationship or a causal relationship.
\end{definition}

\begin{figure}[h]%
\centering
\includegraphics[width=0.9\textwidth]{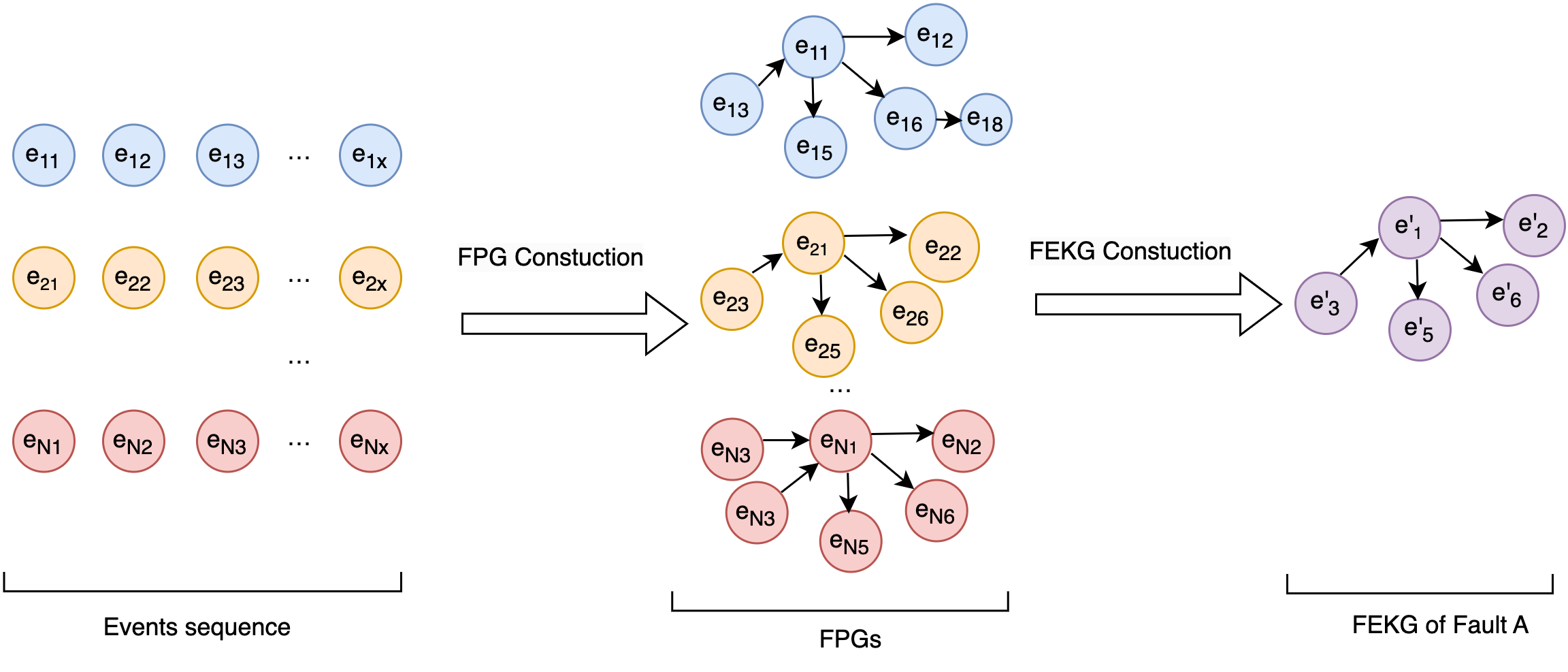}
\caption{An example of transforming the events of fault A to FEKG}\label{Events-FPG-FKG}
\end{figure}

The historical FPGs is a collection of FPG constructed over accumulated faults events. The constructed historical FPGs are used to detect the common structure for FEKG of a fault, shown in Figure \ref{Events-FPG-FKG}. The online FPG is a real time graph constructed when an online fault occurs. This online FPG records the fault propagation relationship graph under the fault scenario. Differently from historical FPGs, online FPG is only one events sequence. The construction of the online FPG includes two steps: 1) \textbf{Event generation}: transferring unstructured data into structured events through data processing module. 2) \textbf{Causal relationship discovery}: using the causal relationship discovery technique for the events generated in step 1 to generate the relationships between the events, we can obtain the online graph within the time window. The constructed online graph will be used to match the most similar knowledge graph constructed for RCA. Figure \ref{Events-FPG-FKG} illustrates transforming the historical events of fault A into FPGs and then FEKG.\\
\indent Two events may be related or unrelated. Among them, in the FPG, the related relationship may be a sequential relationship or a causal relationship. When an event A and event B frequently occur together, it is considered that A and B are related. When judging the relationship between two events, it is not enough to simply determine whether the two events are directly related, but also to consider the environment in which the two events are located, that is, the existing relationship. This is because in some cases, there is no relevant relationship between event A and event B, but event A can cause event B in certain environments. For example, "Setting the POD memory to 10M" cannot directly lead to "POD startup failure", and "setting the POD internal JVM memory to 2G" cannot directly lead to "POD startup failure", but after "setting the POD memory to 10M", "setting the POD internal JVM memory to 2G" will cause "POD startup failure" because the JVM memory is less than the POD memory. The relationship between event A and event B is based on the previous relationship, so when constructing the FPGs, it is necessary to iteratively judge the new relationship.

Algorithm \ref{FPG algorithm} provides a pseudocode representation of how the algorithm is implemented. The process of building the FPG starts with an empty graph. In each iterative step, an event is picked and added to the graph until there are no more events in the sequence, after which the completed graph is returned. Given the strict chronological order of events, it's logical to build the graph in the order of event occurrence. This way, the relevance of events added later can be determined using the events that occurred earlier and their relevance. FPG constructing Algorithm achieves this by using the PopEarliestEvent function to find the earliest event $e_0$ in the event sequence $Q$ during each iteration. If the failure propagation graph is empty at this point, event $e_0$ is directly added to the graph, and the next iteration begins. If not, the GetCandidateGraphs function is used to find all possible candidate graphs that could be formed by adding event  $e_0$ to the FPG. The algorithm then uses the GetBestGraph function to find the candidate graph with the highest score. The RelationClassify function is used to categorize the newly formed relationships in the highest-scoring candidate graph. This graph then becomes the failure propagation graph FPG for this iteration. This process continues until the event sequence $Q$  is empty, at which point the completed failure propagation graph is returned. 
\begin{algorithm}
\caption{$FPG\ Construction$}\label{FPG algorithm}
\begin{algorithmic}[1]
\Statex \textbf{Input:}  $Q:$ the sequence of events captured within a specific fault;
\Statex \hspace{\algorithmicindent}  $M:$ the maximum count of associated events;
\Statex \hspace{\algorithmicindent}  $\theta$: the minimum correlation threshold.
\Statex \textbf{Output:} $FPG$

\State $FPG \gets \{\}$
\State $Q \gets \{e_1, e_2, \ldots, e_N\}$
\While{$Q \neq \emptyset$}
    \State $e_0 \gets \text{PopEarliestEvent}(Q)$
    \If{$FPG == \emptyset$}
        \State $FPG \gets FPG + \{e_0\}$
    \Else
        \State $CandidateGraphs \gets \text{GetCandidateGraphs}(FPG, M, e_0)$
        \State $BestGraph \gets \text{GetBestGraph}(CandidateGraphs, \theta)$
        \If{$BestGraph == \emptyset$}
            \State $FPG \gets FPG + \{e_0\}$
        \Else
        		\For{$(e_i, r, e_j) \in (BestGraph - FPG)$}
    			\State $FPG \gets FPG + \{(e, \phi(e_i, e_j), e_j)\}$
		\EndFor
        \EndIf
    \EndIf
\EndWhile
\end{algorithmic}
\end{algorithm}

\subsection{FEKG Construction}\label{FEKG Construction}
Depending on the assumption that a set of historical data of the same fault type contains all this fault type’s important causal relations, the knowledge graph constructed from a group of same fault type data is reliable. Fault knowledge graph is constructed from historical fault scenarios data. Each FEKG corresponds to a kind of fault case, containing the information(such as root cause types and fault propagation paths) of the specific kind of fault. The following is the definition of the knowledge graph used in this work. 
\begin{definition}Fault Event Knowledge Graph(FEKG).
FEKG is a collection of triples  $(e'_{1}, r, e'_{2})$, where $e'_{1}$ and $e'_{2}$ are abstract events, and r is the relationship between abstract events $e'_{1}$ and $e'_{2}$. r is either a sequential relationship or a causal relationship.
\end{definition}

\indent The knowledge graph is generated by extracting the common fault information of the fault graphs of the same kind of fault. The detailed steps of constructing knowledge graph are shown below: \\
\indent 1) Acquiring event sets $E'_{1}$ , $E'_{2}$, ..., $E'_{n}$ corresponding to all instances $A_{1}$, $A_{2}$, ..., $A_{n}$ of fault type A. For each event set, important events are preserved and connected in chronological order one after another to generate candidate relation sets $R_{1}$, $R_{1}$, ..., $R_{n}$.(A is the fault category, and an instance of fault A refers to fault of A type occurring in a specific time period) \\
\indent 2) Using the trained SVM relation classifiers for candidate relation sets $R_{1}$, $R_{1}$, ..., $R_{n}$ respectively. And only relations judged as "existing" are retained to generate corresponding relation sets $R_1', R_2', \ldots, R_n'$.\\
\indent 3) Combining the relation sets  $R_1', R_2', \ldots, R_n'$ to obtain a fault relation set $R = R_1' \cup R_2' \cup \ldots \cup R_n'$. \\
\indent 4) Constructing knowledge graph of fault type A by merging all relations in relation set R, and replacing the event in the graph with the event type. \\
\indent 5) Domain experts label the root cause type and fault propagation path of a fault in the knowledge graph. \\
Algorithm \ref{FEKG algorithm} demonstrates the construction process from FPGs to FEKG.
\begin{algorithm}
\caption{$FEKG\ Construction$}\label{FEKG algorithm}
\begin{algorithmic}[2]
\Statex \textbf{Input:}  $S\{FPG_1, FPG_2, \dots, FPG_n\}:$ the collection of FPGs for fault $f$ at different time intervals;
\Statex $\mu:$ Inter-cluster linkage threshold.
\Statex \textbf{Output:} $FEKG\{ SG_1, SG_2, \ldots , SG_m \}$

\State $FEKG \gets \{\}$
\State $\Pi \gets GraphClustering(S, \mu)$
\For{each cluster $\Pi_i \in \Pi$}
    \State $SG \gets \{\}$
    \For{each $FPG \in \Pi_i$}
        \State $AbstractFPG \gets Abstract(FPG)$
        \State $SG \gets SG \cap AbstractFPG$
    \EndFor
    \State $FEKG \gets FEKG + SG$
\EndFor

\\
\Statex \#GraphClustering(S, $\mu$)

\State $\pi_0 \gets \{\text{FPG}_1, \text{FPG}_2, \ldots, \text{FPG}_n\}$
\State $k \gets 0$
\While{the number of clusters in $\pi_k$ not equal to 1}
    \State $k \gets k + 1$
    \State $L(P, Q) \gets \text{Calculate the linkage between 2 clusters } P \text{ and } Q$
    \State \text{which have the largest linkage among all pairs of clusters in } $\pi_{k-1}$
    \If{$L(P, Q) < \mu$}
        \State \textbf{break}
    \Else
        \State $R \gets \text{merge the 2 clusters } P \text{ and } Q$
        \State $\pi_k \gets \pi_{k-1} \text{ after replacing } \{P, Q\} \text{ by } R$
        \State $\pi \gets \pi_k$
    \EndIf
\EndWhile

\end{algorithmic}
\end{algorithm}

\subsection{KGroot RCA}\label{KGroot RCA}
\begin{figure}[h]%
\centering
\includegraphics[width=0.4\textwidth]{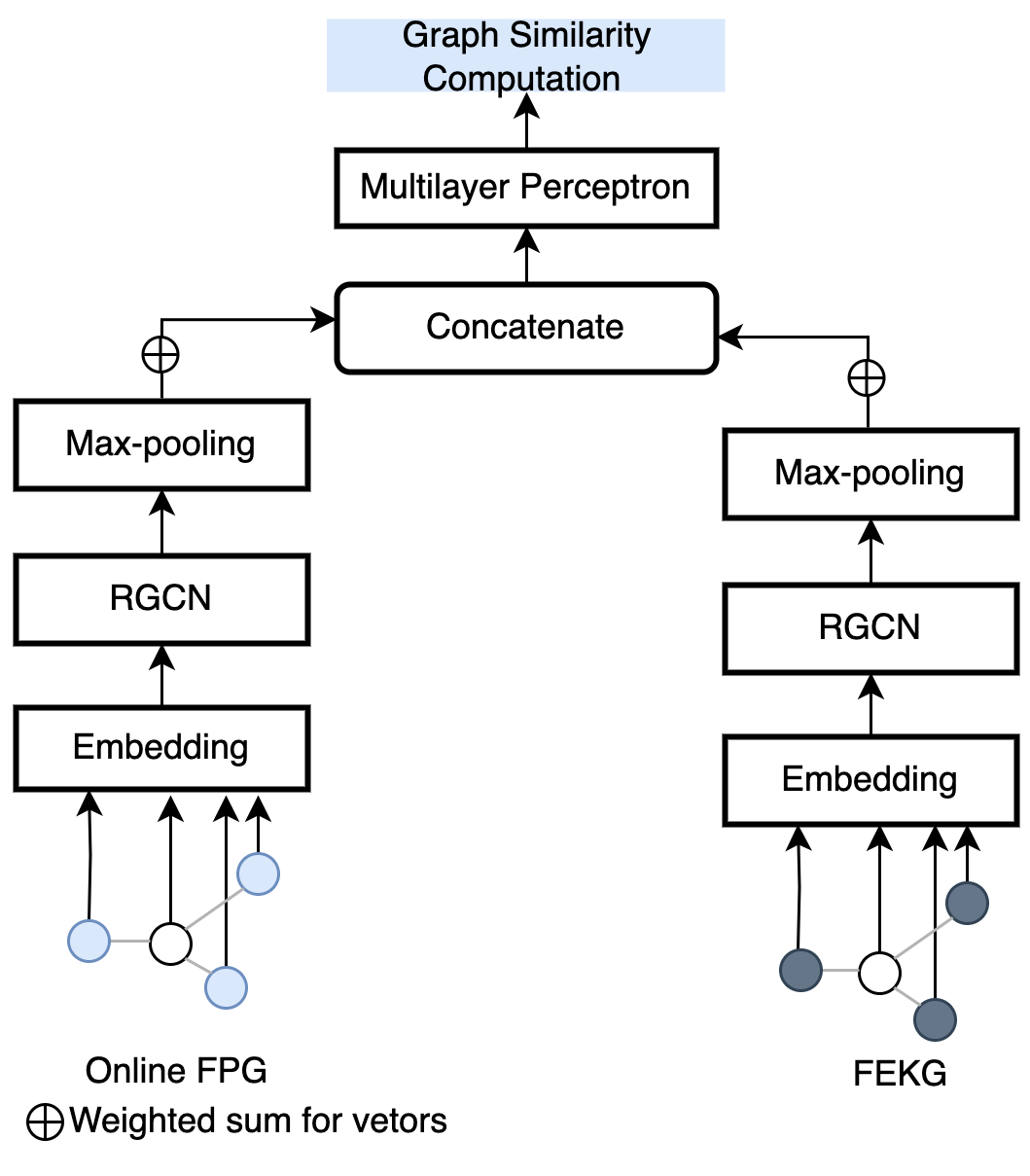}
\caption{Graph similarity computation model}\label{Graph similarity computation model}
\end{figure}

Since the online FPG at the time of failure and the knowledge graph under this fault scenario have the same event logical relationship, they have the highest similarity. KGroot locates possible faults by mapping the online fault graph and FEKG. When a fault occurs, the FPG is first constructed, and then the graph similarity calculation model based on RGCN is used to calculate the similarity between FPG and each FEKG, to find out the FEKG that is most similar to FPG, thereby achieving fault localization.\\
\indent When calculating the similarity between the online FPG and FEKG, the first step is to convert the FPG and FEKG into vector representations and adjacency matrices. The next step uses a relational graph convolution network to extract the features of the graph. This is followed by pooling and concatenation, and finally feeding into a multilayer perceptron to obtain the graph similarity, shown in Figure \ref{Graph similarity computation model}. Graph similarity calculation requires taking into account both node and edge similarities. In the case of online FPG and FEKG, nodes represent events and abstract events respectively. These two different types of nodes can't be directly compared, hence there's a need to standardize the node types. This means that before we can calculate the similarity between the FPG and the FEKG, we need to convert the nodes into abstract events. In this paper we use the word2vec model to obtain a vector representation of events. The word2vec model believes that events with similar contexts are more similar. For example, the context of event $E_x$ is: $E_1$, $E_2$, $E_3$, $E_x$, $E_4$, $E_5$. The context of $E_y$ is: $E_1$, $E_2$, $E_3$, $E_y$, $E_4$, $E_5$. Where $E_i$ are events, $i \in \{1, 2, 3, 4, 5\}$.  The context of $E_x$ is a sequence of events in order before and after $E_x$.  It’s easy to find out $E_x$ and $E_y$ have the same context, and the two events should be similar events.\\
\indent After converting the event into a vector representation that can be processed by the computer, the vector representation of the event can be used as input to the RCA model. Bert model is adopt to convert abstract events into vectors that can be processed by computers. This vector and the adjacency matrix of the graph are input into the relational graph convolutional network to extract the features of the graph. Then pooling and concatenation are performed, and finally the graph similarity is obtained by inputting into a multi-layer perceptron. 
\indent Directed and labeled multigraphs are denoted as $G=(V, E, R)$ with nodes(entities) $v_i \in V$ and labeled edges(relations) $(v_i, r, v_j) \in E$, where $r \in R$ is a relation type. $R$ contains relations both in canonical direction(e.g. cause) and in inverse direction(e.g. cause inv). Specially, we use the initial embedding from word2vec and set window size as 100 to limit the size of graph inputted into RGCN. After acquiring the node representation through RGCN, the maximum pooling layer acts on the graph vector matrix to obtain graph vector representation $g_{\text{{online}}}$ and $g_{\text{{kg}}}$ . $g_{\text{{online}}}, g_{\text{{kg}}} \in \mathbb{R}^{d(\text{{lo}})}$ denotes graph vector representation with the same dimension as the last output layer. Then the final similarity between $g_{\text{{online}}}$ and $g_{\text{{kg}}}$ can be calculated as Equation \ref{pred}.
\begin{equation}\label{pred}
\text{{pred}}_{\text{{online}}}, \text{{kg}} = \text{{softmax}} \left( \sigma \left( W_1 \left( \sigma \left( W_0 \left( \text{{g}}_{\text{{online}}} \oplus \text{{g}}_{\text{{kg}}} \right) + b_0 \right) \right) + b_1 \right) \right)
\end{equation}
where $\oplus$ means \textit{gonline} and \textit{gkg} are concatenated. $W_i$, $b_i$ denote $i$-th linear layer’s weight and bias. The softmax layer will return a vector \textit{predonline,kg} with a dimension of 2, indicating the probability of similarity and the probability of dissimilarity. In order to find the matching one, we only need to confirm the graph in the knowledge base that has the max similarity with the special online graph as the Equation \ref{argmax}.
\begin{equation}\label{argmax}
i = \arg\max_{i \in |K G|} \text{{pred}}_{\text{{online}},i}[1]
\end{equation}

After the above steps, the root cause event type could be choosen. However, in real-time failure scenarios, the root cause event type will correspond to multiple specific events. For example, the CPU overload event type could occur on different ECS. To overcome this problem, the distance and time interval between every candidate event and the alarm event are calculated as Equation  \ref{argmax1}. Following the rule that distance of events is negatively correlated with ranking and the time interval is positively correlated with ranking, the rankings $N_t(\ast)$, $N_d(\ast)$ of event $\ast$ could be calculated. Then the rankings are accumulated by weight to obtain the final ranking of each candidate event, return the top $N$ events in the ranking as the possible root cause events.
\begin{equation}\label{argmax1}
e = \underset{e \in \text{events}}{\mathrm{argmax}} (W_t \cdot N_t(e) + W_d \cdot N_d(e)) \quad
\end{equation}

\section{Experiments}\label{Exp}
In this section, we first introduce the data set used for validation, and then introduce the method of comparative experiment. Finally, we compare the approach of this paper with seven baselines and two degraded versions both in effectiveness and efficiency.
\subsection{Data Set}
In order to test the effectiveness of KGroot, we use an realworld dataset  from a bank system(dataset A) and open-source microservice benchmark system Train Ticket\citep{zhou2018fault} with injected faults(dataset B). The train ticket system is a distributed application of simulating selling train tickets. Following previous works \citep{liu2020unsupervised}\citep{yu2021microrank}, microservice Train-Ticket is deployed on a Kubernetes cluster with 4-node, containing 64 microservices. 156 failures with 23 types are injected by Chaos Mesh\citep{chaos-mesh2022} at random locations. Injected faults include configuration error, network delay, kill pod, CPU overload and memory overload into the system. System metrics include CPU/memory stress on pods/nodes, pod failure, and packet corrupt/loss/delay on pods. The number of faults and the number of fault types in the data set are shown in Table~\ref{dataset} . We also split dataset into training set, validation set, and testing set with proportion of 40\%, 20\%, 40\% respectively. All the evaluation metrics are calculated on the test set. 
\begin{table}[h]
\caption{Dataset summary}\label{dataset}
\centering
\small
\begin{tabular}{p{1.3cm} p{1.6cm} p{1.6cm} p{1.5cm} p{1.5cm} p{3.6cm}}
\hline
\textbf{Dataset} & \textbf{\#Failures} & \textbf{\#Metrics} & \textbf{\#Failure Units} & \textbf{\#Failure Classes} & \textbf{Failure Source} \\ \hline
$A$ & 99 & 2594 & 41 & 41 & Realworld failures of a bank system\\
\hline
$B$ & 156 & 5724 & 1044 & 23 & Injected artificially of open source benchmark\\ \hline
\end{tabular}
\label{your-label}
\end{table}
\begin{table}[!ht]
\centering
\small
\caption{Summary of Baseline Methods}\label{baseline}
\begin{tabular}{p{2.5cm} l p{7cm}}
\hline
\textbf{Classification} & \textbf{Method} & \textbf{Summary of methods} \\
\hline
\multirow{2}{*}{\begin{tabular}{@{}c@{}}Supervised\end{tabular}} & KGroot & Constructing FPG and FEKG based on historical failure, using GCNs to pinpoint the real-time FPG fault type through a ranking strategy. \\
 & D\'ej\`aVu & Using GRU and 1-D CNN for failure feature extractor, GAT for feature aggregator, and classifier suspicious score of each failure. \\
\hline
\multirow{2}{*}{\begin{tabular}{@{}c@{}}Similar\\ Failure Matching\end{tabular}} & JSS & Similar failures identification through graph similarity.  \\

 & iSQUAD& Failure Categorization by root cause labels. \\
\hline
\multirow{4}{*}{\begin{tabular}{@{}c@{}}Traditional\\ ML\end{tabular}} & DT & \\
 & GB &  Traditional ML for classification and regression analysis.\\
 & RF & \\
 & SVM &  \\
\hline
\end{tabular}
\end{table}

\subsection{Baselines}
We compare KGroot with the following state-of-the-art baseline methods, including supervised learning method D\'ej\`aVu\citep{li2022actionable}, similar failure matching methods JSS\citep{brandon2020graph}, iSQUAD\citep{ma2020diagnosing} , and traditional ML methods Decision Tree(DT), Gradient Boosting(GB), Random Forest(RF) and SVM. Similarly, we also conducted an ablation experiment to verify the effect of KGroot under KG and GCN. Table~\ref{baseline} summaries the brief introduction of baselines.

\subsection{Evaluation of Effectiveness}
\begin{table}[!ht]
\centering
\small
\caption{A@K Accuracy of Different Models}\label{accuracy}
\begin{tabularx}{\textwidth}{
  >{\hsize=.9\hsize}X 
  >{\hsize=2.0\hsize}X 
  >{\hsize=1.2\hsize}X 
  >{\hsize=.8\hsize}X
  >{\hsize=.8\hsize}X
  >{\hsize=.8\hsize}X
  >{\hsize=.8\hsize}X
  >{\hsize=.45\hsize}X 
}
\hline
\textbf{Dataset} & \textbf{Classification} & \textbf{Method} & \textbf{A@1} & \textbf{A@2} & \textbf{A@3} & \textbf{A@5} & \textbf{MAR} \\
\hline
\multirow{11}{*}{A} & Supervised & D\'ej\`aVu & 69.47\% & 71.05\% & 81.57\% & 82.13\% & 3.51\\
\cline{2-8}
& \multirow{2}{*}{\parbox{4cm}{Similar \\ Failure Matching}} & JSS & 31.57\% & 55.26\% & 65.78\% & 70.14\% & 26.34\\
& & iSQUAD& 10.52\% & 34.21\% & 63.15\% & 71.23\% & 4.01\\
\cline{2-8}
& \multirow{4}{*}{Traditional ML} & DT & 37.86\% & 43.87\% & 44.32\% & 51.27\% & 11.21\\
& & GB & 44.73\% & 65.78\% & 73.68\% & 78.63\% & 3.10 \\
& & RF & 63.15\% & 73.68\% & 81.57\% & 80.84\% & 2.48 \\
& & SVM & 28.94\% & 50\% & 60.52\% & 76.31\% & 3.76 \\
\cline{2-8}
& \multirow{3}{*}{Ablation Study} & KGroot & \textbf{71.24\%} & \textbf{73.29\%} & \textbf{84.27\%} & \textbf{85.28\%} & \textbf{3.84} \\
& & KGroot w/o KG & 59.32\% & 69.14\% & 79.51\% & 81.06\% & 3.72 \\
& & KGroot w/o GCN & 68.13\% & 70.13\% & 82.16\% & 83.18\% & 4.38 \\
\hline
\multirow{11}{*}{B} & Supervised & D\'ej\`aVu & 74.37\% & 85.93\% & 90.62\% & 83.61\% & 17.75 \\
\cline{2-8}
& \multirow{2}{*}{\parbox{4cm}{Similar \\ Failure Matching}} & JSS & 9.37\% & 9.37\% & 15.62\% & 21.87\% & 280.99 \\
& & ISQ& 23.43\% & 31.25\% & 32.81\% & 32.81\% & 267.58 \\
\cline{2-8}
& \multirow{4}{*}{Traditional ML} & DT & 65.62\% & 73.43\% & 73.43\% & 73.43\% & 132.12 \\
& & GB & 65.62\% & 75\% & 75\% & 76.62\% & 26.26 \\
& & RF & 73.75\% & 73.75\% & 73.75\% & 83.6\% & 4.53 \\
& & SVM & 6.25\% & 7.812\% & 7.812\% & 15.62\% & 95.35 \\
\cline{2-8}
& \multirow{3}{*}{Ablation Study} & KGroot & \textbf{75.18\%} & \textbf{86.12\%} & \textbf{93.5\%} & \textbf{85.17\%} & \textbf{10.75} \\
& & KGroot w/o KG & 73.21\% & 83.45\% & 90.15\% & 81.69\% & 11.81 \\
& & KGroot w/o GCN & 71.89\% & 84.51\% & 89.16\% & 80.73\% & 9.71 \\
\hline
\end{tabularx}
\end{table}
Following related works\citep{li2022actionable}, top N accuracy(A@K) and mean average rank(MAR) are introduced to evaluate the effectiveness of related methods. A@K represents the probability that the actual result is hit among the K possible results given by the fault localization. MAR is the mean of the average suggested rank of all ground truths of each failure. It represents how many recommendations should be checked to diagnose a failure on average, and smaller is better. Precision, Recall, and F1 Score are chosen to evaluate the performance of different algorithms.\\
\begin{table}[!ht]
\centering
\small
\caption{Precision, Recall and F1 Evaluation} \label{precision}
\begin{tabular}{cllccc}
\hline
\textbf{Dataset} & \textbf{Classification} & \textbf{Method} & \textbf{Precision} & \textbf{Recall} & \textbf{F1}\\
\hline
\multirow{11}{*}{A} & Supervised & D\'ej\`aVu & 72.38\% & 71.23\% & 74.23\% \\
\cline{2-6}
& \multirow{2}{*}{\parbox{3cm}{Similar \\ Failure Matching}} & JSS & 31.57\% & 31.57\% & 19.94\% \\
& & iSQUAD& 15.78\% & 14.28\% & 4.5\% \\
\cline{2-6}
& \multirow{4}{*}{Traditional ML} & DT & 37.86\% & 37.86\% & 37.86\% \\
& & GB & 55.26\% & 50\% & 52.5\% \\
& & RF & 68.42\% & 61.9\% & 65\% \\
& & SVM & 39.47\% & 35.71\% & 28.19\% \\
\cline{2-6}
& \multirow{3}{*}{Ablation Study} & KGroot & \textbf{75.41\%} & \textbf{73.24\%} & \textbf{76.18\%} \\ 
& & KGroot w/o KG & 70.17\% & 69.18\% & 73.17\% \\
& & KGroot w/o GCN & 73.12\% & 70.31\% & 75.19\% \\
\hline
\multirow{11}{*}{B} & Supervised & D\'ej\`aVu & 74.37\% & 74.37\% & 74.37\% \\
\cline{2-6}
& \multirow{2}{*}{\parbox{3cm}{Similar \\ Failure Matching}} & JSS & 9.37\% & 9.37\% & 9.37\% \\
& & ISQ& 23.43\% & 23.43\% & 23.43\% \\
\cline{2-6}
& \multirow{4}{*}{Traditional ML} & DT & 65.62\% & 65.62\% & 65.62\% \\
& & GB & 65.62\% & 65.62\% & 65.62\% \\
& & RF & 73.75\% & 73.75\% & 73.75\% \\
& & SVM & 6.25\% & 6.25\% & 0.78\% \\
\cline{2-6}
& \multirow{3}{*}{Ablation Study} & KGroot & \textbf{76.31\%} & \textbf{72.5\%} & \textbf{74.39\%} \\ 
& & KGroot w/o KG & 74.31\% & 71.34\% & 72.15\% \\
& & KGroot w/o GCN & 72.89\% & 69.18\% & 70.73\% \\
\hline
\end{tabular}
\end{table}

\begin{table}[!h]
\caption{Efficiency Evaluation}\label{Efficiency}
\begin{tabular}{@{}cllc@{}}
\hline
\textbf{Dataset} & \textbf{Classification} & \textbf{Method} & \textbf{Time(ms)} \\
\hline
\multirow{9}{*}{A} & Supervised & D\'ej\`aVu  & 349 \\
\cmidrule{2-4}
& \multirow{2}{*}{Similar Failure Matching} & JSS & 629 \\
 &  & iSQUAD& 638 \\
\cmidrule{2-4}
& \multirow{4}{*}{Traditional ML} & DT & 613 \\
 &  & GB & 527 \\
 &  & RF & 512 \\
 &  & SVM & 503 \\
\cmidrule{2-4}
& \multirow{3}{*}{Ablation Study} & KGroot & \textbf{351} \\
 &  & KGroot w/o KG & 338 \\
 &  & KGroot w/o GCN & 343 \\
 \hline
 \multirow{9}{*}{B} & Supervised & D\'ej\`aVu  & 597 \\
\cmidrule{2-4}
& \multirow{2}{*}{Similar Failure Matching} & JSS & 885 \\
 &  & ISQ& 891 \\
\cmidrule{2-4}
& \multirow{4}{*}{Traditional ML} & DT & 817 \\
 &  & GB & 734 \\
 &  & RF & 758 \\
 &  & SVM & 746 \\
\cmidrule{2-4}
& \multirow{3}{*}{Ablation Study} & KGroot & \textbf{578} \\
 &  & KGroot w/o KG & 561 \\
 &  & KGroot w/o GCN & 557 \\
\hline
\end{tabular}
\end{table}

\indent We compare KGroot with 7 baselines and 2 degraded versions over dataset A and dataset B repectively . The 7 baselines are implemented to verify our proposal's superiority over existing solutions. GCN and Knowledge graphs importance are evaluated separately by these 2 degraded versions. Table~\ref{accuracy} presents the RCA results. Each experiment is repeated ten times and presents the average results. KGroot always keeps the best top-k accuracy. On dataset B, KGroot achieved the highest accuracy in predicting A@1, A@2, A@3, and A@5, which are 75.18\%, 86.12\%, 93.5\%, and 85.17\% respectively. KGroot outperforms D\'eja\`Vu by 2.88\% in A@3 accuracy. MAR of KGroot  is 10.75 and outperform the most baselines by 39.43\%$\sim$96.17\%(except RF).\\
In our experiments, we also investigate KGroot and various models in their precision, recall and F1 scores. From the results(Table~\ref{precision}), the KGroot model performed the best among all models. It has a precision of 76.31\%, a recall of 72.5\%, and an F1 score of 74.39\% on dataset B. Compared with other models, KGroot performed excellently in precision, which means it surpassed other models in predicting positive classes. At the same time, although its recall rate is slightly lower than D\'ej\`aVu and RF, its F1 score(the harmonic mean of precision and recall) is still the highest, indicating that KGroot achieved a good balance between maintaining prediction accuracy and coverage. Overall, the experimental results show that the KGroot model outperforms overall performance in this experiment, with high precision and recall, and the highest F1 score both in dataset A and B. 
\subsection{Evaluation of Efficiency}
We compared the prediction efficiency with the baseline method on a server equipped with Gen intel core i7 and GeForce RTX 3090. As shown in Table~\ref{Efficiency}, KGroot can locate the cause of the fault within one second, and compared with manual positioning, KGroot achieves second-level automatic fault diagnosis. This means that when developers and operation and maintenance personnel receive monitoring alarms, they also know the possible causes of the fault. This has met the requirements for fault location in the production environment. Compared with other baseline methods, KGroot has excellent fault location time performance. \\
\section{Conclusion and Future works}\label{Con}
 \label{sec:Con}
This paper proposed an actionable faults localization approach, KGroot. Motivated by the frequent occurrence of repeated faults, we adopted a supervised model to learn how to pinpoint faults from a vast amount of historical failure data. We designed a novel deep learning-based model based on a knowledge graph and RGCN, which achieved good performance for this limited scenario. Learning from offline historical faults, when a system failure occurred, KGroot automatically constructed an online graph according to the fault event. After computing online graph similarity with historical faults knowledge graph in RGCN, KGroot chose the most similar possible root causes for online faults and made recommendations. By comparative experimenting on public data, KGroot achieved a 93.5\% accuracy rate for the top 3 possible root cause locations, and achieved 70\%--80\% traceability accuracy. This paper also compared the other seven baselines and two degraded versions of the approach, and the experimental results proved that our approach demonstrated practical performance.\\
\indent While the RCA method proposed in this study has seen improvements in accuracy and performance over existing methods, there's still potential for further enhancement. The method of tracing origins in this paper is done by comparing the similarity between the real-time fault propagation graph and the fault knowledge graph, which overlooks the inherent characteristics of the propagation map. Future studies could delve into the properties of fault propagation and attempt to incorporate the fault knowledge graph for inferring propagation pathways. In consideration of the real-world scenario, it is vital to establishing a positive cycle learning from past failures, getting positive feedback from users and learning from feedback for accurate faults diagnosis. 
Going forward, we could consider utilizing the fault knowledge graph for fault prediction, automatic inspection, and automatic fault handling, to tap into its full potential and construct a more comprehensive and robust end-to-end intelligent operation and maintenance system. 
\section*{Acknowledgments}\label{Ack}
This work is supported by the Natural Science Foundation of China (Grant No. U21A20488). We thank the Big Data Computing Center of Southeast University for providing the facility support on the numerical calculations in this paper. 


\bibliographystyle{elsarticle-harv} 
\bibliography{KGroot}






\end{document}